\documentclass[journal]{IEEEtran}
\usepackage{graphicx}
\usepackage{subcaption}
\usepackage{amsmath}
\usepackage{cite}
\usepackage{floatrow}
\usepackage{gensymb}
\usepackage{amsfonts}
\usepackage{etoolbox}
\usepackage{enumerate}
\usepackage{multirow}
\usepackage{romannum}
\usepackage{booktabs}
\usepackage{graphicx}
\usepackage{balance}
\usepackage{siunitx}
\usepackage{algorithmic}
\usepackage[linesnumbered,ruled,vlined]{algorithm2e}
\usepackage{soul}
\usepackage{lipsum}
\usepackage{graphicx}
\usepackage{stfloats}
\usepackage [autostyle, english = american]{csquotes}
\usepackage{amsmath}

\usepackage{amsmath}
\usepackage{csquotes}
\usepackage{ mathrsfs }

\MakeOuterQuote{"}
\usepackage{caption}
\usepackage[hidelinks,colorlinks=true,filecolor=black,linkcolor=green,citecolor=blue,urlcolor=black] {hyperref}
\def\BibTeX{{\rm B\kern-.05em{\sc i\kern-.025em b}\kern-.08em
    T\kern-.1667em\lower.7ex\hbox{E}\kern-.125emX}}
\usepackage{hyperref}

\ifCLASSINFOpdf
\else
\fi
\begin{document}
%
\title{STEF-DHNet: Spatiotemporal External Factors Based Deep Hybrid Network for Enhanced Long-Term Taxi Demand Prediction}
%
%
%

\author{Sheraz Hassan, Muhammad Tahir, Momin Uppal, Zubair Khalid, Ivan Gorban, and Selim Turki}
\maketitle

\begin{abstract}
Accurately predicting the demand for ride-hailing services can result in significant benefits such as more effective surge pricing strategies, improved driver positioning, and enhanced customer service. By understanding the demand fluctuations, companies can anticipate and respond to consumer requirements more efficiently, leading to increased efficiency and revenue. However, forecasting demand in a particular region can be challenging, as it is influenced by several external factors, such as time of day, weather conditions, and location. Thus, understanding and evaluating these factors is essential for predicting consumer behavior and adapting to their needs effectively. Grid-based deep learning approaches have proven effective in predicting regional taxi demand. However, these models have limitations in integrating external factors in their spatiotemporal complexity and maintaining high accuracy over extended time horizons without continuous retraining, which makes them less suitable for practical and commercial applications. To address these limitations, this paper introduces STEF-DHNet, a demand prediction model that combines Convolutional Neural Network (CNN) and Long Short-Term Memory (LSTM) to integrate external features as spatiotemporal information and capture their influence on ride-hailing demand. The proposed model is evaluated using a long-term performance metric called the rolling error, which assesses its ability to maintain high accuracy over long periods without retraining. The results show that STEF-DHNet outperforms existing state-of-the-art methods on three diverse datasets, demonstrating its potential for practical use in real-world scenarios.
\end{abstract}


%
\IEEEpeerreviewmaketitle

\section{Introduction}
%
%
%
%
\IEEEPARstart{T}{axi} demand prediction is a critical aspect of effective urban management, which plays a vital role in ensuring the livability and sustainability of our cities. Urban management involves coordinating various services and resources to meet the needs of a growing population. Transportation is a crucial component of urban management, and predicting taxi demand is essential to optimize resource utilization, enhance citizens' transportation experience, and support ride-hailing companies' fleet management, pricing strategies, and marketing initiatives. Effective and accurate demand forecasts can create sustainable cities, improve citizens' quality of life, increase economic productivity, and reduce environmental impact \cite{yigitcanlar2015planning, aljoufie2011urban}.

\begin{figure}[ht]
    \centering
        \includegraphics[scale=.355]{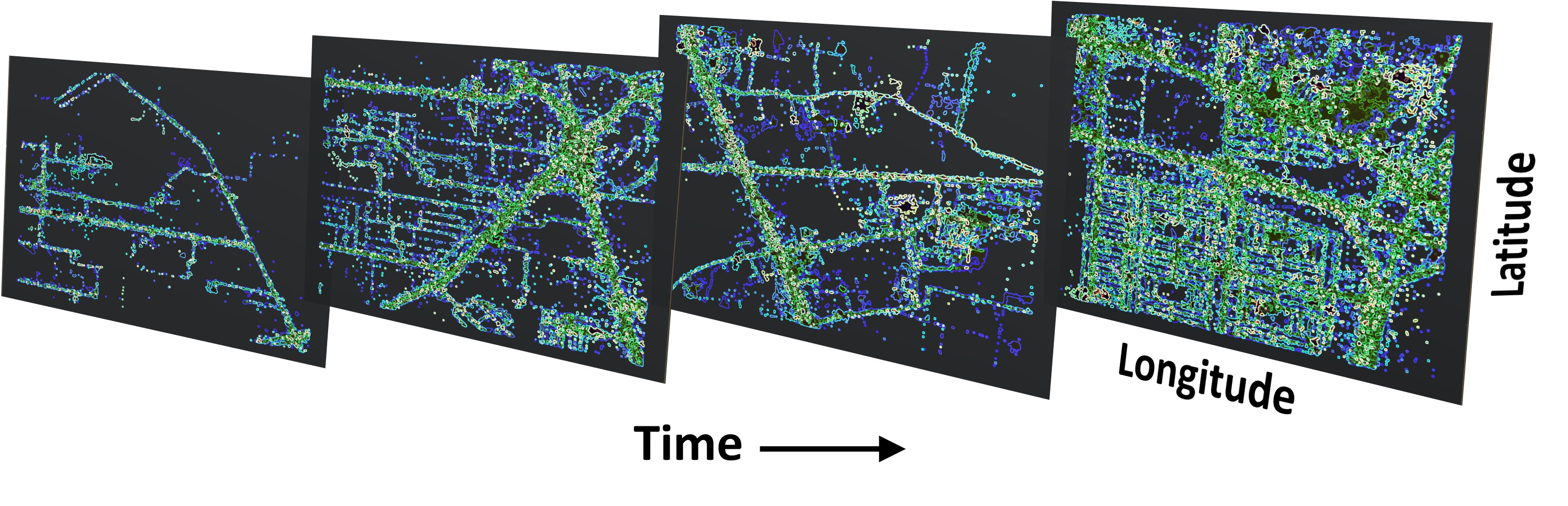}
    \caption{Illustrating the Complexities of Spatiotemporal Dependencies in Taxi Demand Forecasting}
    \label{fig:Spatiotemporal}
\end{figure}

The prediction of taxi demand presents a challenge due to the ever-changing spatiotemporal nature of traffic data, as depicted in Fig. \ref{fig:Spatiotemporal}, and its intricate dynamic dependencies \cite{yin2021deep}. Apart from other factors, the complexities arise from the interplay between regions. The demand of a particular region can be influenced by its neighboring regions and simultaneously be correlated with regions that share a similar context despite being distant. Additionally, there are non-linear dependencies among different time periods, with the prediction at a specific time being correlated with historical observations from previous hours, days, or even weeks \cite{geng2019spatiotemporal}.

However, in addition to the complex spatial and temporal relationships, demand for ride-hailing services is influenced by various external factors, such as weather conditions, traffic patterns, and even political events. These factors can vary in their impact on different regions and can change over time. Research has shown that weather conditions may impact ride-hailing services more so than traditional taxis, a phenomenon that may be attributed to the adaptive pricing structure of the former \cite{etminani2019individuals}. Furthermore, areas with a higher concentration of workplaces relative to residences, such as central business districts, have been found to experience higher demand for ride-hailing services \cite{qian2017taxi,yang2018analysis}. Therefore, to accurately predict demand, a representation of spatiotemporal data that takes into account these dynamic externalities and the unique characteristics of different regions is essential.

In recent years, traditional Machine Learning (ML) models such as Support Vector Machines (SVM), Gradient Boosting Machines (GBM) \cite{lin2020leveraging} and its modification Extreme Gradient Boosting (XGBoost) \cite{vanichrujee2018taxi} have been used for ride-hailing demand prediction due to the complexity of the spatiotemporal data involved. However, with the increasing use of deep learning techniques, attention has shifted towards utilizing these models to enhance accuracy and efficiency. Deep learning models have the ability to capture more complex and non-linear patterns in spatiotemporal data, leading to more accurate and reliable predictions. One category of deep learning models that have gained widespread adoption in ride-hailing demand prediction is graph-based models such as Adaptive Graph Convolutional Recurrent Networks (AGCRN) \cite{bai2020adaptive}. Graph Neural Networks (GNN) are employed in these models to capture the intricate spatial interconnections between different regions, making them more effective in predicting demand. GNNs are neural models that reflect the relationships between graphs through inter-node message transmission in the graph \cite{zhou2020graph}. Another model class that has been put forth is grid-based models, which has its own advantages \cite{liu2017grid}. These models utilize a grid architecture to depict the relationships between spatiotemporal variables and commonly employ CNN to model the spatial dependencies between various time increments. Many grid-based models such as Spatial-Temporal Dynamic Network (STDN) \cite{yao2019revisiting} and Deep Multi-View Spatial-Temporal Network (DMVST-Net) \cite{yao2018deep} have demonstrated efficacy in capturing the complex spatiotemporal patterns in the data and have been employed in numerous studies to predict taxi demand, pick-up volumes, and inflow/outflow trends.

In this paper, we present a novel grid-based deep learning model for predicting demand for ride-hailing services. The intricacies posed by the complexity of the data necessitate the consideration of both temporal and spatial dependencies in our model. To effectively model the interactions between regions, the spatial dependencies are captured using CNN layers. In contrast, the temporal dependencies, which involve the non-linear relationship between current predictions and past observations, are captured through the deployment of Long Short-Term Memory (LSTM) layers. Our model also effectively incorporates external factors in their true complexity as a spatiotemporal representation of the data, which can vary in effect across regions and change over time, leading to enhanced demand forecasting. The efficacy of the proposed model is demonstrated through its evaluation on multiple datasets, showcasing its proficiency in capturing both the spatiotemporal dependencies and external factors for improved demand forecasting. Additionally, a new performance metric has been used to evaluate demand prediction models in real-world scenarios, taking long-term predictions without retraining into account. The proposed model outperforms the current state-of-the-art methods in terms of testing error and this metric. The main contributions of the work can be summarized as follows:

\begin{enumerate}
    \item We propose a novel grid-based deep learning model that leverages the strength of CNN and LSTM layers to predict the demand for ride-hailing services. The model is designed to capture both spatial and temporal dependencies of the demand data and effectively incorporate external factors in its true complexity. Our model provides a more accurate representation of real-world scenarios by considering the actual spatiotemporal intricacy of the external factors.
    \item We evaluate the efficacy of the model through comparative analysis with multiple state-of-the-art alternatives on diverse datasets, including two publicly available datasets from New York City and one dataset from an emerging country. The results demonstrate the remarkable accuracy and robustness of our proposed model in demand prediction.
    \item We assess our model and other grid-based state-of-the-art models' accuracy in real-world applications using a newly introduced performance metric, called rolling error. This metric considers the model's previous output as input to generate the subsequent outputs, making it an efficient way to evaluate the accuracy of a model over an extended period. Our proposed model outperforms the state-of-the-art methods on this metric as well, demonstrating its ability to generate accurate demand predictions without the need of continuous retraining.
\end{enumerate}
The paper is organized as follows. Section II gives an overview of the existing literature. In Section III, the methodology of our proposed approach is discussed in detail. Section IV provides the results we have achieved. Section V concludes our study, summarizing the essential findings and discussing the potential avenues for further research and development in this field.

\begin{figure*}[ht!]
    \centering
        \includegraphics[scale=.395]{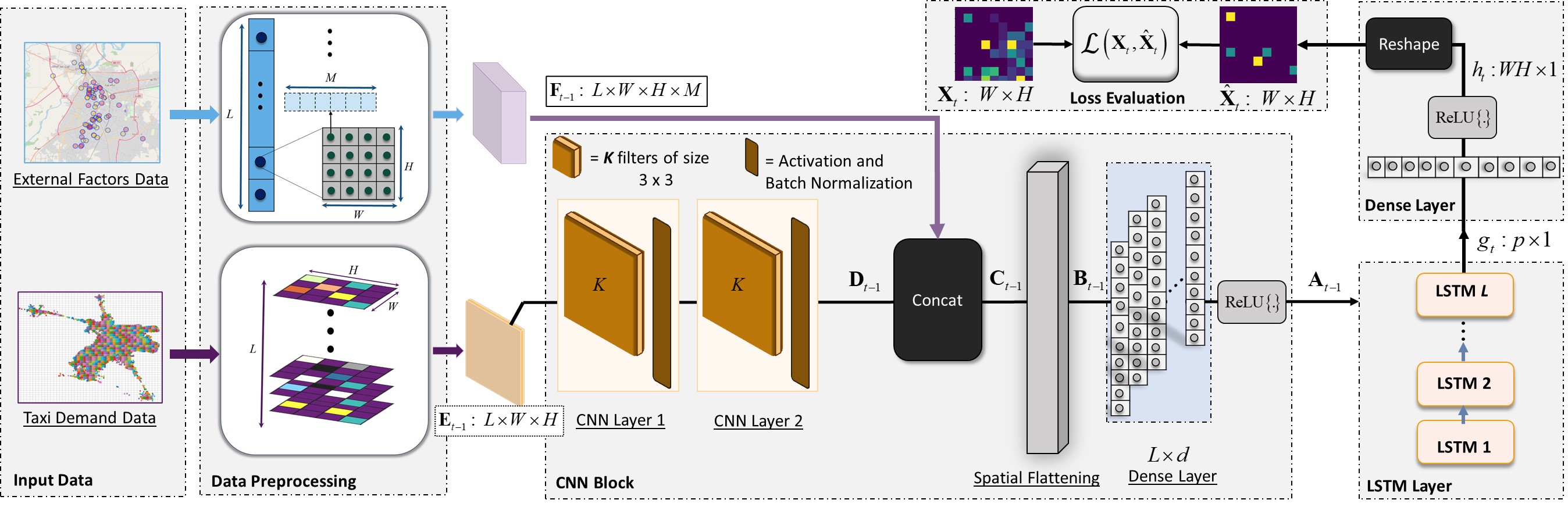}
    \caption{Visualizing the Methodology: A Comprehensive Flowchart of Data Preprocessing, Model Structure, and Outcome}
    \label{fig:flowchart}
\end{figure*}
\section{Related Work}

The demand for accurate taxi predictions has grown significantly in recent years, inspiring a plethora of research in the field \cite{zhao2020unifying, liu2020building}. Various traditional as well as machine learning techniques have been applied to this task, including ARIMA \cite{jamil2017taxi}, XGBoost \cite{vanichrujee2018taxi}, and Random Forest \cite{rajendran2021predicting}. With the proliferation of deep learning techniques, researchers have increasingly turned to these methods to address the challenges of predicting taxi demand. These deep learning approaches can be categorized into two main domains: grid-based and graph-based. A comprehensive comparison of these techniques is available in a survey paper by Jiang et al. \cite{jiang2021dl}. 

Advancements in graph based deep learning methods have enabled the development of diverse approaches to this problem \cite{zeng2021robust}. For instance, Zhao et al. proposed T-GCN, a sophisticated model that combines Gated Recurrent Unit (GRU) and GNN to predict traffic flow in urban areas \cite{zhao2019t}. Another example is the hybrid model presented by Karingula et al. \cite{karingula2022boosted}. This model combines the strengths of CatBoost and Neural Networks (NN) to tackle the spatiotemporal complexity inherent in time series forecasting for taxi demand prediction. Graph WaveNet \cite{wu2019graph} adopts an adaptive graph rather than a static one. Huang et al. went further by introducing the DSTGNN model that effectively handles complex spatiotemporal dependencies in taxi demand predictions \cite{huang2022dynamical}. As our approach to predicting taxi demand relies on a grid-based deep learning method, this section will concentrate on reviewing the current literature related to the application of these techniques.

 Liao et al., in their study \cite{liao2018large}, showed that Deep Neural Networks (DNNs) based on a grid system can achieve superior results compared to traditional machine learning techniques. He emphasized the importance of carefully designing a suitable DNN architecture and established the foundation for using DNNs to analyze complex spatiotemporal data. Ziat et al. \cite{ziat2017spatio} proposed a grid-based model that leveraged latent representations and Recurrent Neural Networks (RNNs) to capture spatiotemporal dynamics. However, this approach could not account for periodic patterns in the data. Koutnik et al. \cite{koutnik2014clockwork} attempted to address the challenge of modeling long-term dependencies through their Clockwork RNN (CW-RNN) model. However, it was inadequate for handling patterns spanning over days or weeks where the input had high dimensional features. Several other studies \cite{lv2014traffic,zhang2017understanding} focused on traffic density prediction using deep learning but disregarded either spatial, temporal, or both correlations. Zhang et al. \cite{zhang2017deep} used a deep residual network for crowd flow prediction, but the model was limited to convolutional layers and could not effectively capture temporal correlations. To improve on these limitations, Zonoozi et al. proposed the Periodic-CRN (PCRN) method \cite{zonoozi2018periodic}, which adapted a Convolutional Recurrent Network (CRN) to capture spatial and temporal correlations accurately. It also learns and incorporates explicit periodic representations. Similarly, Yao et al. proposed the DMVST-Net framework \cite{yao2018deep} that uses a temporal view to capture correlations between future demand and near-time points using LSTM, a spatial perspective to understand the local spatial correlations through a local CNN, and a semantic view to identify correlations among regions that exhibit similar temporal patterns. However, this approach has limitations, as it can only work for a small region due to its use of a local CNN. It also fails to incorporate external factors as spatiotemporal data into the model, which are critical components for accurate demand prediction.
 
One of the possible ways to improve the accuracy of taxi demand prediction is to introduce point-of-interests (POI) data intelligently into the  model structure. Building on this idea, Cao et al. \cite{cao2021bert} incorporated POIs using their functional similarity score in a BERT-based Deep Spatial-Temporal Network (BDSTN) model. Their model exploits the complex spatial-temporal relations from heterogeneous local and global features. However, a limitation of the BDSTN model is that it ignores the influence of temporal variations of other key external factors that can significantly impact travel behavior. Lin et al. developed the DeepSTN+ model \cite{lin2019deepstn+}, which considers the impact of POI distributions, time, and location attributes on taxi demand prediction by utilizing a fully-connected layer in the ConvPlus block. Although the model captures long-range spatial dependence, its complexity with a large number of parameters leads to slow training times, high computational costs hindering its practical use in real-world applications. The high number of trainable parameters also negatively impacts its performance in the long run.

\section{Methodology}
In this section, we start by defining the traffic forecasting challenge utilizing grid models. Then, we outline the steps we take to preprocess our data so that it can serve as input for the model. Our goal is to incorporate relevant spatiotemporal external factors in the process. Finally, we introduce our novel approach to tackle the complexity of spatiotemporal external factors in our model. A clear illustration of the entire methodology is provided in Fig. \ref{fig:flowchart}.

\subsection{Problem Definition}

The demand prediction problem of ride-hailing services refers to forecasting the number of ride requests in a specific geographic area over time. The prediction problem is highly non-trivial as the demand variable is influenced by a number of  spatiotemporal external factors such as the time of day, weather conditions, public transportation accessibility, and location itself. We divide a geographical region into total $N$ spatial grids with $N=WH$. We use variables $w$ and $h$ for indexing these spatial regions as $w=1, 2, \cdots, W$ and $h=1, 2, \cdots, H$. The input demand sequence in a particular spatial region is defined as $\{ X_t^{w,h} \in \mathbb{Z}^+\}$ where $t$ denotes the time instant whose range is denoted as $t=1, 2, \cdots, T$ with $T$ being the total time.  When considered for the complete grid, we denote it as $ \mathbf{X}_t$ of size $ W \times H $. The input sequence used to  make a demand prediction at time $t$ is formed by concatenating \textit{the most recent} $L$ values of  $\mathbf{X}_t $ as $\mathbf{E}_{t-1} = \left[ \mathbf{X}_{t-1}; \mathbf{X}_{t-2}; \cdots ; \mathbf{X}_{t-L+1} \right]$ where $\mathbf{E}_{t-1} $ is of shape $ L \times W \times H $ where $L$ is time-window lag/width. 

As already explained, a significant contribution of the proposed STEF-DHnet model is to intelligently introduce and combine external spatiotemporal factors which impact the demand value. We consider total $M$ external spatiotemporal factors indexed as $m=1,2,\cdots, M$ containing temporal and spatial weights (binary) corresponding to each location grid. We can arrange weights corresponding to these factors in a vector $\mathbf{y} $ of size $M \times 1$. Stacking these into spatial grids produces $\mathbf{Y}_t $ of size $W \times H \times M$ tensor for a given time instant $t$ where $t=1,2,\cdots,T$. We can now extract \textit{the most recent} $L$ values of $\mathbf{Y}_t $ as well into a new shape $\mathbf{F}_{t-1} = \left[ \mathbf{Y}_{t-1}; \mathbf{Y}_{t-2}; \cdots ; \mathbf{Y}_{t-L+1} \right]$ where $\mathbf{F}_{t-1} \in L \times W \times H \times M$. The goal of the model is now to produce a prediction of  $ \mathbf{X}_t $, denoted as $ \hat{\mathbf{X}}_t $ using $\mathbf{E}_{t-1}$ and $\mathbf{F}_{t-1}$ as
\begin{equation}
   \hat{\mathbf{X}}_t = \textit{STEF-DHNet}\left( \mathbf{E}_{t-1} , \mathbf{F}_{t-1} | {\mathbf{\Theta}}\right)
\end{equation}
where ${\mathbf{\Theta}}$ denotes the parameter vector of the proposed model.

 

\subsection{Data Preprocessing}
The data preparation process involves the processing of input  demand data and external factors data. For each spatial grid, we first establish its boundaries, followed by the extraction of external factors with appropriate  temporal weights, transforming them into spatiotemporal data. A detailed description of the process is provided below and depicted in Fig. \ref{fig:flowchart} for visual representation.

\subsubsection{Demand Data}
Once the boundary of the demand data grid is established, it is divided into $N$ equal regions of shape $W \times H$, where $H$ represents the height, and $W$ represents the grid's width. For temporal disintegration of the demand data, we set the  temporal resolution $t$  to 1 hour, resulting in a total of $T$ hours of demand data. The model requires demand data with a lag of $L$ i.e., in order to predict the demand at hour $t$, the model requires the demand data from previous $L$ hours. As a result, we stack $L$ number of consecutive grids  to form the input for the model, $\mathbf{E}_{t-1} $. This process of stacking grids effectively incorporates the prior time-based dependencies into the data, making it suitable to be used as input to the model.

\subsubsection{External Factors} 
Incorporating external factors into the demand prediction model requires careful data extraction and preprocessing. We extracted $M$ different external factors such as bus stations, universities, colleges, commercial hubs, and airports. Their spatial location is known to impact consumer demand for ride-hailing services in that particular region \cite{almunawar2021customer}. The location data of these external factors were obtained through geocoding techniques, converting addresses or names into geographic coordinates and then assigning them to specific grid cells within the geographic area. This allows the model to consider the proximity of these external factors to the regions of the grids and their impact on demand in those locations \cite{zhao2016predicting}. For each region in the grid, $M$ external factors are considered. To account for the varying impact that these external factors have on demand throughout the time, binary weights are assigned based on observed demand variation. For instance, high demand are observed at airports after flight arrivals and at universities during class hours in the morning and evening. This results in an external factor spatiotemporal grid of dimensions $ W \times H $ with $M$ external factors. To meet the requirement of our model, which takes input with a lag of $L$, $L$ number of grids are stacked together to form the model's final input data $\mathbf{F}_{t-1}$. The complete process is illustrated in Fig. \ref{fig:flowchart} for visual understanding.
\subsection{Proposed STEF-DHNet Model}
Our proposed STEF-DHNet model merges the advantages of CNN and LSTM layers to predict the demand for ride-hailing services while also considering the spatiotemporal nature of external factors such as universities and commercial hubs while fusing them within the network architecture. We use a time lag of $L=4$ hours  to predict the demand for the next hour. The first two layers of the network are convolutional layers. We denote  $f_i \left( . | \mathbf{\Theta}_i \right)$ to represent the operation (including batch-normalization) performed by $i-$th layer of the network with parameters vector $\mathbf{\Theta}_i$ to represent
\begin{equation}
   \mathbf{D}_{t-1} ={f}_2\left( {f}_1\left( \mathbf{E}_{t-1} | \mathbf{\Theta}_1 \right) | \mathbf{\Theta}_2 \right)
\end{equation}
For each layer, we use $K=32$  kernels with kernel size $3 \times 3$. The output of these two layers, $\mathbf{D}_{t-1}$, is a tensor of size $L \times W \times H \times K$. To include external factors, we concatenate $\mathbf{D}_{t-1}$ with the external factors tensor $\mathbf{F}_{t-1}$ along the last axis, creating a unified representation of the grid and time information along with the  external factors as
\begin{equation}
   \mathbf{C}_{t-1}= \textit{CONCAT}\left( \mathbf{D}_{t-1} , \mathbf{F}_{t-1} \right)
\end{equation}
The resulting  tensor $\mathbf{C}_{t-1}$ is of size  $L \times W \times H \times (K+M)$, where $M$ is the number of external factors. Note that we use $M$ number of external factors for every region. In the next step, only the spatial dimensions of $\mathbf{C}_{t-1}$ are flattened to a new shape of $L \times WH(K+M)$, denoted by $\mathbf{B}_{t-1}$. 
\begin{equation}
   \mathbf{B}_{t-1}= \textit{FLATTEN}\left( \mathbf{C}_{t-1} \right)
\end{equation}
Before passing the data to LSTM layer, we use $L$ fully-connected dense layers (FC-D), each with an output of size $d$, to generate a dense output for each time lag. The resulting shape of size $L \times d$ is denoted by $\mathbf{A}_{t-1}$. This operation is written as
\begin{equation}
    \mathbf{A}_{t-1} = \textit{FC-D} \left( \mathbf{B}_{t-1} | \mathbf{\Theta}_{D_1} \right)
\end{equation}
where $\mathbf{\Theta}_{D_1}$ denotes the parameter vector for the dense layer.   The LSTM layer follows the dense layers and aims to capture temporal patterns in the data. 
\begin{equation}
   {g}_{t} = \textit{LSTM}\left( \mathbf{A}_{t-1} | \mathbf{\Theta}_L \right)
\end{equation}
where $\mathbf{\Theta}_{L}$ denotes the parameter vector for the LSTM layer. The output of the LSTM layer, denoted by $\mathbf{g}_{t}$, is fed into another FC-D with size $N$ to produce an output 
\begin{equation}
    {h}_{t} = \textit{FC-D} \left( g_{t} | \mathbf{\Theta}_{D_2} \right)
\end{equation}
where $\mathbf{\Theta}_{D_2}$ denotes the parameter vector for this second dense layer. This final dense layer is then reshaped into a tensor of dimensions $W \times H$, effectively yielding the predicted demand $ \hat{\mathbf{X}}_t $, taking into consideration both the spatiotemporal demand information, and external factors as 
\begin{equation}
   \hat{\mathbf{X}}_{t-1}= \textit{RESHAPE}\left( {h}_{t} \right)
\end{equation}
The proposed STEF-DHNet model is characterized by the following parameter vector $\mathbf{\Theta} = \left[ \mathbf{\Theta}_{1}, \mathbf{\Theta}_{2}, \mathbf{\Theta}_{D_1}, \mathbf{\Theta}_{L}, \mathbf{\Theta}_{D_2}  \right]$. 

\subsection{Loss Function and  Model Training}
In order to train STEF-DHNet, we use mean-absolute error (MAE) loss function  defined as 

 \begin{align}
     \mathcal{L}\left( \mathbf{X}_{t}, \hat{\mathbf{X}}_{t} \right) &= Mean\left(\vert \mathbf{X}_{t} - \hat{\mathbf{X}}_{t}\vert \right) \\
     &= \frac{1}{N}\sum_{w=1}^{W}\sum_{h=1}^{H}\vert  X_t^{w,h} - \hat{X_{t}}^{w,h} \vert 
 \end{align}
Once we have calculated the loss, we use backpropagation to train the model. During backpropagation, the gradients of the loss function with respect to the weights are computed. The gradients are then used to update the weights using an ADAM optimizer. By iteratively minimizing the loss function, our model gradually learns to make more accurate predictions.

\section{Experiments and Results}
In this section, we carry out experiments on different datasets to compare the performance of the  proposed STEF-DHNet method with state-of-the-art baselines for taxi demand prediction problem. In our experiments, we utilized the Tensorflow framework on an NVIDIA RTX 3090 GPU with 24 GB of memory. The optimization of the parameters of the neural network was performed using the Adam optimization method with a constant learning rate of 0.001 and a batch size of 256. To ensure optimal performance, we used early-stop in all the experiments. The early-stop round and the max epoch were set to 100 and 2000 in the experiment, respectively.

\subsection{Performance Metrics}
We have used  Mean Absolute Error (MAE), Root Mean Squared Error (RMSE) of each time averaged over all regions, and Mean Absolute Percentage Error (MAPE) as our evaluation metrics for performance comparison. These metrics are listed below. 
\begin{align}
    MAE &= \frac{1}{NT} \sum_{w=1}^{W}\sum_{h=1}^{H} \sum_{t=1}^{T} \left\vert  X_t^{w,h} - \hat{X_{t}}^{w,h} \right\vert  \\
    RMSE &= \sqrt{ \frac{1}{NT} \sum_{w=1}^{W}\sum_{h=1}^{H} \sum_{t=1}^{T} \left( X_t^{w,h} - \hat{X_{t}}^{w,h} \right)^2 }  \\
    MAPE &= \frac{1}{T}  \sum_{t=1}^{T} \left( \frac{\left\vert  \mathbf{X}_t - \hat{\mathbf{X}_t} \right\vert }{\mathbf{X}_t}  \right) \times 100
\end{align}

The metrics mentioned above are evaluated for three data sets at one-hour intervals. 

\subsection{Data Sets}
In our study, we assess our method by conducting experiments on three distinct data sets related to the prediction of taxi demand. Two of these data sets, NYCTAXI and NYCBIKE, are publicly accessible and commonly used in the literature for taxi demand prediction problem. The third data set, XYZCAREEM, is sourced from Careem, a ride-hailing service. Here is a brief overview of each data set:
\begin{enumerate}
    \item NYCTAXI \cite{nycopendata}: This publicly available data set provides information on taxi rides in New York City. We used data from January 2016 to February 2016 and divided it into 900 small regions. To convert the data into a grid format, we mapped the regions onto a grid with 30$\times$30 dimensions.
    \item NYCBIKE \cite{bucketloading}: This data set provides information about bike sharing in New York City and is also publicly accessible. We used data from January 2022 to March 2022 and defined 400 regions with equal latitude and longitude boundaries. We then aligned all these regions to form a grid with 20$\times$20 dimensions.
    \item XYZCAREEM: This data set is proprietary and contains information on Careem demand in a region of Pakistan. The data includes pick-up and drop-off locations, timestamps, and trip distances and is available for a period of 3 months. We divided the data into 285 small regions spatially, which we then mapped onto a grid of size 15$\times$19.    
\end{enumerate}


\subsection{Baselines}
We compared our proposed method to several deep learning models that utilize grid-based techniques. These models are considered to be state-of-the-art baselines for predicting taxi demand and forecasting traffic flow. The list includes the following. 
\begin{enumerate}
\item ST-ResNet \cite{zhang2017deep}: ST-ResNet is a CNN-based deep learning model that leverages the unique properties of spatiotemporal data, including temporal closeness, period, and trend, to predict the inflow and outflow of crowds in a specific region of a city.
\item DeepSTN+ \cite{lin2019deepstn+}: DeepSTN+ is a convolutional deep learning model designed for forecasting taxi inflow and outflow in a city.
\item DMVST-Net \cite{yao2018deep}: DMVST-Net utilizes three views to capture the complex spatiotemporal relationships within the data, improving the accuracy of demand predictions. The temporal view captures time-based patterns and correlations, the spatial view captures local spatial correlations, and the semantic view captures correlations among regions that share similar temporal patterns.
\item Periodic-CRN \cite{zonoozi2018periodic}: PCRN utilizes convolutional and recurrent layers to capture spatiotemporal correlations within the data.
\end{enumerate}


\subsection{Model training, testing, and rolling}
We split the total available data samples for each dataset $\mathcal{D}$ into three non-overlapping parts $\mathcal{D}_{train}$, $\mathcal{D}_{test}$, $\mathcal{D}_{roll}$, where $\mathcal{D}_{train}$ is used to train/validate the models, $\mathcal{D}_{test}$ is used to test the goodness of fit of every model on unseen test data, and $\mathcal{D}_{roll}$ is used to roll (as explained below) the models for a given number of time. For every respective part of the data split, we obtain corresponding training, testing and rolling errors (MAE, RMSE, and MAPE) for all methods and on all datasets. 

 Unlike traditional performance metrics that rely on a model's predictions at one specific time, the rolling error accounts for the accumulation of errors over multiple predictions over an extended time duration. The term \enquote{rolling} refers to the metric being calculated using a rolling window approach, where the prediction for time $t$  is used as input to generate the next input for the model which then generates the subsequent prediction and so on.  The rolling error metric provides valuable insights into the accuracy of a model over an extended period and helps in selecting the best model that minimizes the error over time, without the need for frequent retraining. 

The split is performed as: 65\% for training, 15\% for validation, and 20\% for testing (including rolling). To ensure consistency and comparability, we adjusted each data set to a temporal resolution of one hour. We use the last week as the evaluation data for rolling predictions, corresponding to 168 hours of data. 



\subsection{ Results and Discussion}
We conducted an evaluation of various cutting-edge models based on their training, testing, and rolling error metrics and compared their performance against our proposed STEF-DHNet model. The results are presented in Table Table \ref{table:Error}. It is clear  that DeepSTN+ performed well on the NYCTAXI and NYCBIKE datasets, but it produced less accurate results on the XYZCAREEM dataset due to its unique challenges as an emerging country dataset. On the other hand, our proposed STEF-DHNet model demonstrated superior performance in terms of training and testing error metrics across almost all datasets. We have presented the performance evaluation results of all models on the testing, training, and rolling error metrics in Table \ref{table:Error}.

As we can see from Table \ref{table:Error}, the rolling error metric is a crucial indicator of a model's ability to predict accurately over an extended period. Our proposed STEF-DHNet model outperformed other models on this metric as well, demonstrating its effectiveness for long-term predictions. Although P-CRN also exhibited promising results in terms of the rolling error metric, our proposed model consistently performed better across all datasets, including the larger NYCTAXI dataset. 


\begin{table*} \centering
\begin{small}
\begin{tabular}{@{}l|l|rrr|rrr|rrr@{}}\toprule
 \textbf{Dataset} & \textbf{ Method}  & \textbf{} & \textbf{Training}& \textbf{} & \textbf{} & \textbf{Testing} & \textbf{} & \textbf{} & \textbf{Rolling}& \textbf{} \\ \midrule
\midrule

  \textbf{}&\textbf{}& \textbf{MAE} & \textbf{RMSE} & \textbf{MAPE} & \textbf{MAE} & \textbf{RMSE} & \textbf{MAPE} & \textbf{MAE} & \textbf{RMSE} & \textbf{MAPE}\\ \midrule
\textbf{} &\textbf{STEF-DHNet} & \textbf{02.25} & \textbf{05.21} & \textbf{04.79}
 & \textbf{02.51} & \textbf{06.03} & \textbf{13.52} & \textbf{02.61} & \textbf{06.39} & 17.51\\ 
\textbf{} &\textbf{Periodic-CRN} & 02.54 & 06.04 & 15.15
 & 02.62 & 06.53 & 16.80 & 02.62 & 06.54 & \textbf{16.33}\\ 
\textbf{XYZCAREEM} &\textbf{DMVST-Net} & 02.96 & 05.86 & 23.84 & 02.97 & 06.08 & 18.12 & 18.89 & 34.68 & 527.4\\ 
\textbf{} &\textbf{DeepSTN+} & 02.92 & 05.71 & 20.20 & 02.98 & 06.05 & 16.04 & 03.50 & 07.14 & 26.48\\ 
\textbf{} &\textbf{ST-ResNet} & 02.97 & 05.82 & 17.51 & 03.15 & 06.48 &  16.52 & 03.40 & 07.38 & 29.55\\ \midrule
\midrule

\textbf{} &\textbf{STEF-DHNet} & 01.11 & 12.55 & \textbf{02.53}
 & 02.34 & 27.40 & 09.00 & \textbf{02.33} & \textbf{26.44} & \textbf{09.97}\\ 
\textbf{} &\textbf{Periodic-CRN} & 03.43 & 38.72 & 1508
 & 03.07 & 28.94 & 10.22 & 03.05 & 28.51 & 10.54\\ 
\textbf{NYCTAXI} &\textbf{DMVST-Net} & \textbf{00.96} & \textbf{07.36} & 04.24
 & \textbf{01.44} & \textbf{12.25} & \textbf{03.56} & 18.32 & 185.7 & 98.60\\ 
\textbf{} &\textbf{DeepSTN+} & 02.90 & 20.33 & 131.4 & 03.04 & 21.25 & 10.88 & 04.97 & 41.16 & 23.26\\ 
\textbf{} &\textbf{ST-ResNet} & 11.49 & 30.74 &
 206.4 & 12.26 & 33.83 & 19.99 & 1429 & 2567 & 11775\\ 
\midrule
\midrule

\textbf{} &\textbf{STEF-DHNet} & \textbf{00.72} & \textbf{02.00} & \textbf{05.61}
 & \textbf{01.45} & \textbf{04.59} & 15.63 & \textbf{01.45} & \textbf{04.35} & \textbf{21.82}\\ 
\textbf{} &\textbf{Periodic-CRN} & 01.53 & 05.25 & 54.01 & 02.77 & 09.67 & 43.38 & 02.14 & 07.33 & 37.25\\ 
\textbf{NYCBIKE} &\textbf{DMVST-Net} & 01.00 & 02.55 & 15.92 & 01.58 & 04.63 & \textbf{13.54} & 07.05 & 23.76 & 321.2\\ 
\textbf{} &\textbf{DeepSTN+} & 01.03 & 02.42 & 21.71 & 01.87 & 07.56 & 18.40 & 02.47 & 07.75 & 41.21\\ 
\textbf{} &\textbf{ST-ResNet} & 01.22 & 02.75 & 23.48 & 01.90 & 04.50 & 17.97 & 03.77 & 12.76 & 94.96 \\ 
\midrule
\bottomrule
\end{tabular}
\end{small}
\caption{Performance comparison of different models in terms of training, testing, and rolling error}
\label{table:Error}
\end{table*}

\subsection{Efficiency Evaluation}
\begin{figure*}[t]
    \centering
        \includegraphics[scale=.3]{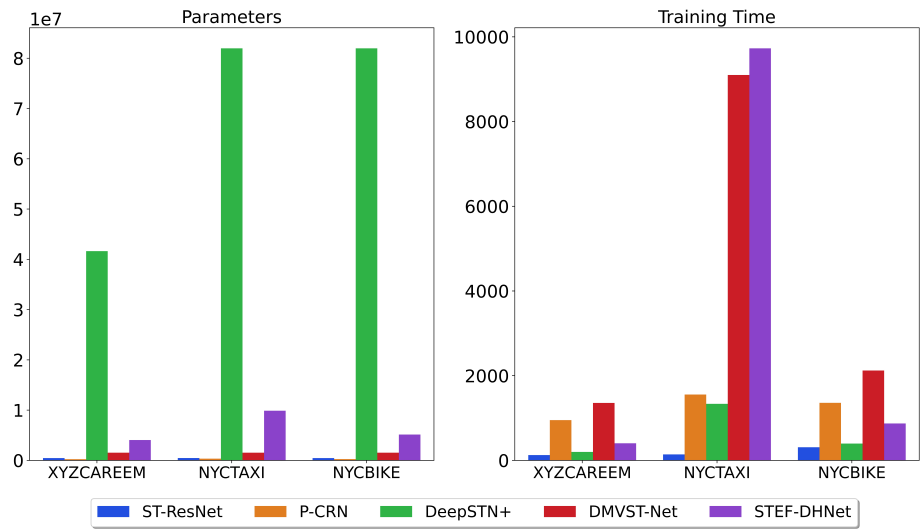}
    \caption{Number of parameters and training time for each model}
    \label{fig:efficiency}
\end{figure*}

When evaluating the performance of the models, it's important to consider not only their prediction accuracy but also their model complexity and training time. To this end, we present the results of model complexity (number of parameters) and  time required to train the model in Fig. \ref{fig:efficiency}. It is clear that the DeepSTN+ has the highest number of parameters, making it the most complex model. Conversely, ST-ResNet has the lowest number of parameters for each dataset, making it the simplest model. Meanwhile, DMVST-Net had the highest training time, and ST-ResNet had the lowest training time.

Our proposed STEF-DHNet model has the second-highest number of parameters and training time, although the difference is considerable compared to DeepSTN+ and DMVST-Net. It's important to note that the high number of parameters in our model is due to the unique nature of the spatiotemporal data being used. Fig. \ref{fig:efficiency}  shows the efficiency evaluation results for all the models evaluated, illustrating the balance between model complexity and training time for each model. Our proposed STEF-DHNet model demonstrates excellent performance in terms of model complexity and training time, making it a practical and viable option for spatiotemporal predictions.


\section{Conclusion}
We  present a grid-based deep learning model, called STEF-DHNet, for ride-hailing companies demand prediction. The model is based on a novel deep learning architecture  that combines CNN and LSTM layers, allowing it to successfully capture the impact of external factors on demand. The proposed model has been shown to outperform the existing state-of-the-art methods in short-term as well as long-term predictions while being computationally efficient. This work has the potential to bring tangible benefits to ride-hailing companies by allowing them to make more informed decisions and respond to consumer demand more efficiently. 


%



\ifCLASSOPTIONcaptionsoff
  \newpage
\fi



%

\bibliographystyle{IEEEtran}
\bibliography{Paper}
\end{document}